\documentclass[10pt,twocolumn,letterpaper]{article}

\usepackage{graphicx}
\usepackage{times}
\usepackage{epsfig}
\usepackage{amsmath}
\usepackage{amssymb}
\usepackage[hyphens]{url}
\usepackage[draft]{hyperref}
\usepackage{balance}
\usepackage{color}
\usepackage{graphicx}
\usepackage{booktabs}
\usepackage{multirow}
\usepackage{colortbl}

\usepackage[draft]{hyperref}

\providecommand{\keywords}[1]{\textbf{\textit{Index terms---}} #1}

\begin{document}

\title{\textit{It All Matters:}\\ Reporting Accuracy, Inference Time and Power Consumption \\for Face Emotion Recognition on Embedded Systems}
\date{}

\author{Jelena Milosevic\\
Institute of Telecommunications, TU Wien\\
{\tt\small jelena.milosevic@tuwien.ac.at}
\and
Dexmont Pe\~{n}a\\
Movidius an Intel Company \\
{\tt\small dexmont.pena@intel.com}
\and
Andrew Forembsky\\
Movidius an Intel Company\\
{\tt\small andrew.forembsky2@mail.dcu.ie}
\and
David Moloney\\
Movidius an Intel Company \\
{\tt\small david.moloney@intel.com}
\and
Miroslaw Malek\\
ALaRI, Faculty of Informatics, USI\\
{\tt\small miroslaw.malek@usi.ch}
}

\maketitle

\begin{abstract}
While several approaches to face emotion recognition task are proposed in literature, none of them reports on power consumption nor inference time required to run the system in an embedded environment. Without adequate knowledge about these factors it is not clear whether we are actually able to provide accurate face emotion recognition in the embedded environment or not, and if not, how far we are from making it feasible and what are the biggest bottlenecks we face.

The main goal of this paper is to answer these questions and to convey the message that instead of reporting only detection accuracy also power consumption and inference time should be reported as real usability of the proposed systems and their adoption in human computer interaction strongly depends on it. 
In this paper, we identify the state-of-the art face emotion recognition methods that are potentially suitable for embedded environment and the most frequently used datasets for this task.
Our study shows that most of the performed experiments use datasets with posed expressions or in a particular experimental setup with special conditions for image collection. Since our goal is to evaluate the performance of the identified promising methods in the realistic scenario, we collect a new dataset with non-exaggerated emotions and we use it, in addition to the publicly available datasets, for the evaluation of detection accuracy, power consumption and inference time on three frequently used embedded devices with different computational capabilities. 
Our results show that grey images are still more suitable for embedded environment than color ones and that for most of the analyzed systems either inference time or energy consumption or both are limiting factor for their adoption in real-life embedded applications.

\end{abstract}


\section{Introduction}
 
Face emotion recognition has wide range of applications and an automatic system that would be able to accurately recognize human emotions would be a giant step towards improvements of human and computer interactions. 
Promising candidates for accurate face emotion recognition systems are deep learning based methods, since, due to the increased amount of available data, these methods start to outperform, in terms of detection accuracy, many other commonly used machine learning algorithms.
Due to this, deep learning, and more precisely Convolutional Neural Networks (CNNs) are a desirable algorithm to run in order to achieve good detection results.
One of the main requirements for face emotion recognition systems, apart from accuracy, is their suitability for run-time usage and low-power consumption. 
CNNs, that provide good accuracy on one side, on the other are computationally complex and their usage on low power devices is a challenging task. 
However, although power consumption is one of the main criteria for the developed system to be used in practice, it is our observation that it is frequently neglected. Same situation is observed also with inference time. More precisely, we have noticed that none of the face emotion recognition methods proposed in the literature, reports its power consumption and inference time, and without them, it is difficult to understand if the proposed method is suitable for run-time scenario or not.

We aim at facilitating the adoption of face emotion recognition methods in real world scenarios. 
In order to achieve this, we first survey state-of-the-art face emotion recognition methods and then implement those that are potentially suitable for embedded devices and report their detection performance on a dataset composed of three different datasets (publicly available CKPlus and Jaffe datasets and a Custom dataset collected by us) and power consumption and inference time on three different embedded devices: Movidius Neural Computing Stick (NCS); Raspberry Pi; and Intel Joule. 
In this way, we want to evaluate how suitable the proposed methods are for the embedded environment, identify potential bottlenecks and support the comparison of future work in this domain with the existing methods from all three different aspects that matter: accuracy, power consumption and inference time.  
Having in mind the aforementioned, the main contributions of this paper are as follows:
\begin{enumerate}
    \item Listing and brief assessment of the available datasets frequently used in face emotion recognition (Section \ref{sec:datasets})
    \item Review of the state-of-the-art methods using CNN in face emotion recognition and selection of the most suitable candidates for embedded systems environment (Section \ref{sec:relatedwork})
    \item Collection of the dataset representative of realistic scenarios and non-exaggerated emotions to be used for the evaluation of the methods in addition to the publicly available datasets (Section \ref{sec:dataset})
    
    \item Analysis and evaluation of power consumption, inference time and detection accuracy of the most promising state-of-the-art algorithms on the collected dataset and discussion on observed research challenges and obstacles in the deployment of the face emotion recognition methods in embedded environment (Section \ref{sec:results})
\end{enumerate} 

\section{Frequently Used Datasets}
\label{sec:datasets}
In the analysis that we perform, we first summarize the list of datasets frequently used by researchers in the face emotion recognition domain. 
Following, the results of this analysis. 

\begin{itemize}
\item \textbf{Cohn-Kanade AU-Coded Facial Expression} has two versions of which the second one, named CKPlus, is the most frequently used among researchers. 
In the initial version 1,486 sequences from 97 posers are included. Each sequence starts with a neutral expression and proceeds to a peak expression. 
Images are labeled with the emotion label that refers to what expression was requested. Detailed description of CK can be seen at \cite{ck}.
In the second version, posed and non-posed (spontaneous) expressions are included. For posed expressions, the number of sequences is increased from the initial release by 22\% and the number of subjects by 27\%. As with the initial release, the target expression for each sequence is fully labelled. 
More details on the dataset are given in \cite{ckplus}. 

\item \textbf{Japanese Female Facial Expression (JAFFE)} database contains 213 grey images of all 7 facial expressions posed by 10 Japanese female models. 
Each image has been rated on 6 emotion adjectives by 60 Japanese subjects. More information about the dataset can be found at \cite{jaffe1} and \cite{jaffe2}.                    
\item \textbf{The CMU Multi-PIE} database contains more than 750,000 images of 337 people recorded in up to four sessions over the span of five months. Subjects were imaged from 15 view points and under 19 illumination conditions while displaying a range of facial expressions. In addition, high resolution frontal images were acquired as well. In total, the database contains more than 305 GB of face data. More information about the database, including its collection and structure, so as performed baseline experiments, can be found in \cite{pie1} and \cite{pie2}.

\item \textbf{Static Facial Expression in the Wild (SFEW)} is the dataset released in 2015 for the purpose of the classification of the facial emotions on static images challenge. The baseline challenge detection accuracy was 35.93\%. 
It contains images from $95$ subjects and expressions belonging to all seven basic emotion categories collected in real environment, as opposed to JAFFE, CKPlus and MultiPIE that are created in lab environment.
More information about it can be found at \cite{sfew}.

\item \textbf{Karolinska Directed Emotional Faces (KDEF)} dataset contains 4900 pictures of human facial expressions of emotion, that were initially intended for use in medical research.
The set contains 70 individuals, each displaying 7 different emotional expressions, each expression being photographed (twice) from 5 different angles \cite{kdef}.  

\item \textbf{Facial Expression Recognition Challenge (FER2013)} dataset consists of 48x48 pixel grayscale images of faces, where the faces have been automatically registered so that the face is more or less centered and occupies about the same amount of space in each image. The task is to categorize each face based on the emotion shown in the facial expression in to one of seven categories.
More information about the dataset, the FER2013 challenge and its winners can be found at \cite{fer2013}. 

\item \textbf{MMI Facial Expression} database consists of over 2900 videos and high-resolution still images of 75 subjects. It is fully annotated for the presence of action units (AUs) in videos, and partially coded on frame-level, indicating for each frame whether an AU is in either the neutral, onset, apex or offset phase. More details on this dataset can be found at \cite{mmi}.

\item \textbf{A spontaneous facial action intensity database (DISFA)} described in \cite{disfa} is one of the rear databases that contains images of spontaneous expressions of $27$ subjects, recorded while watching video clips. This database is not emotion-specific encoded, but rather contains a list of 12 AUs annotation where each of AUs is expressed on the scale from zero to five.

\item \textbf{BP4D} dataset is a part of facial expression recognition and analysis challenge \textbf{(FERA)} proposed in \cite{fera}. It contains AUs of young persons when they were responding to the emotion-triggering tasks. The dataset is split into training part, that consist images from 41 subjects and test, with images from 20 subjects. 

\end{itemize}

\section{Related work} 
\label{sec:relatedwork}
In last years, CNNs were shown to outperform many other state-of-the-art methods in the areas related to image classification \cite{levi2015}.
For this reason, we focus on the existing face emotion recognition methods based on CNNs and, in this Section, give an overview of their accuracy as reported by the authors and the datasets used. 

In \cite{song2014} the authors propose a facial expression recognition approach running on a smartphone. This approach is based on client-server architecture, i. e. the picture is taken with the phone and then it is sent to an external server to perform its classification. The proposed network consists of four convolutional layers and one fully connected layer. The result of the classification is one of five emotions: anger, happiness, sadness, surprise and neutral.  The method is evaluated on four different datasets (one publicly available, two collected by the authors, and one collected from Internet by downloading images with corresponding emotions). The highest obtained detection accuracy was 99.2\% on the publicly available CKPlus dataset. 

In \cite{yu2015} a face emotion recognition model based on the ensemble classification is used. It first performs face detection using the ensemble of three state-of-the-art models: joint cascade detection and alignment; Deep-CNN-based detector; and Mixtures of Trees. Then it uses an ensemble of multiple deep CNNs and based on their weighted output makes a decision on the emotion. This architecture based on an ensemble of CNNs achieves 61.29\% detection accuracy on SFEW2.0 dataset. 

Ghosh et al. \cite{ghosh2015} propose a multi-label CNN approach to learn a shared representation between multiple action units and directly from the input image (no engineered features). The experiments are performed on individual datasets (CKPlus, DISFA, and BP4D) and by using cross-dataset.
The deployed CNN architecture consists first, of a normalization layer and then, two convolution layers followed by pooling. Then, two fully connected layers are used, followed by a classifier of action units.   
The reported detection results are of 84.6\% on DISFA dataset and 75.8\% on BP4D dataset.   

Mollahosseini et al. \cite{Mollahosseini2015} present a DNN learning architecture focused on the examination of the network ability to perform cross-database classification while training on databases that have limited scope, and are often specialized for a few expressions. They use seven publicly available facial expression databases: MultiPIE, MMI, CKPlus, DISFA, FERA, SFEW, and FER2013.
The obtained results outperform the state of the art for some databases and show to be in pair of the state-of-the-art for the remaining datasets. 

In \cite{levi2015} an approach using Local Binary Patterns (LBP) as pre-processing is used. Then a deep CNN model is trained to use these LBP to perform face emotion recognition. Using SFEW dataset the authors achieve an accuracy of 54.56\%. 
The source code of the paper is released by the authors and is available online at \url{http://www.openu.ac.il/home/hassner/projects/cnn_emotions/}. Together with the source code necessary to replicate the pre-processing stage, the authors also released the python notebook for example usage, so as trained CNN models.

The approach proposed in \cite{Kim2015} trained multiple deep CNNs by varying network architectures, input normalization, and weight initialization as well as by adopting several learning strategies to use large external databases. The proposed approach was the winner of the SFEW challenge with obtained maximal accuracy of 61.6\% using SFEW dataset. 

In DeXpression paper \cite{DeXpression2016}, another architecture for emotion face recognition based on CNN is proposed. The datasets used are CKplus and MMI Facial Expression Database and the obtained accuracy achieved on them is 99.6\% and 98.63\%, respectively.  

The work presented in \cite{Alizadeh2016} proposes a face emotion detection model based on CNN using FER2013 dataset. Two models, shallow and deep, are created and compared in the paper, in terms of the accuracy on seven different emotions, loss function and convergence speed. In summary, validation accuracy was increased by 18.46\% using the deep network in most cases, but for some emotions (surprise and fear) it actually decreased the accuracy. Additionally, the authors tested the system adding additional potentially meaningful features to it, together with the once used by CNN already, and confirmed that the CNN was able to use the significant features by itself already and the added features did not help much in improving accuracy. The obtained accuracy is given per emotion and in the case of shallow models the obtained maximum is 75\% for happy faces and in the case of the deep model the maximum is again on the detection of happy emotion and it is 80.5\%.  

Duncan et al. \cite{duncan2016} propose a human emotion, real-time detection system able to work in different scenarios, angles and lighting conditions. The authors created an application called HappyNet which displays the detected emotion over a human face. 
The authors used three different datasets:    the   extended  Cohn-Kanade   dataset (CKPlus), the Japanese Female Facial Expression  (JAFFE)  database, and a homebrew database, created by authors using data from five individuals. The detection accuracy on JAFFE 62\%, on CKPlus 90.7\%, and on homebrew database 90.9\% on training dataset and 57\% on testing dataset (results reflect the accuracy under testing with perfect conditions: perfect lighting, camera at eye level, subject facing camera with an exaggerated expression). 
If these conditions are not fulfilled, the accuracy drops significantly.
The solution is developed for python and can be found on github at  \url{https://github.com/GautamShine/emotion-conv-net}. The authors release code necessary to replicate HappyNet, to retrain HappyNet on new data, and to generate a new training set. 
  
Two CNN based architectures are explored in \cite{Ruiz-Garcia2016}: the first one explores the impact of reducing the number of deep learning layers
and the second one uses of a novel image representation approach that splits the input images and makes use of two deep learning streams. The authors used the Karolinska directed Emotional faces database (KDEF). The obtained results show that both network architectures perform the best on the detection of the happy faces, while the most frequently misclassified are the neutral faces.

In FaceNet2ExpNet \cite{facenet2016} two stage training algorithm is proposed. It first trains the convolutional layers of the expression net, regularized by the face net, and then, in the refining stage, it appends fully-connected layers to the pretrained convolutional layers and train the whole network jointly. Accuracy results obtained for the datasets CKPlus, Oulu-CASIA, TFD, and SFEW are: 96.8\%, 87.71\%, 88.9\%, 48.19\% respectively.

In \cite{Laranjeira2017} another approach to face emotion recognition is proposed, that using CNN and CKPlus achieves detection accuracy of 90\%. 

In addition to surveyed research papers, also nViso network used to perform emotion recognition was found available. It is a 12-layer network, developed as a part of the Eyes of Things project and trained by the company nViso. As reported by its designers, the network is designed to use a small amount of memory and it takes only 0.9MB \cite{Deniz2017}. 

In Table \ref{tab:relatedwork} we summarize the aforementioned papers, by outlining their detection accuracy, used dataset and if the open source code for them is released or not. In case where the authors did not name their proposed network, the names of the first authors of the corresponding papers were used.   

\begin{table*}
\centering

\begin{tabular}{|l|c|l|c|}
\hline
\textbf{Reference} & \textbf{Accuracy} & \textbf{Dataset} & \textbf{Open Source} \\ \hline
Song \cite{song2014} & \begin{tabular}{l} 99.2\% \\ 97.1\% \\ 95.5\% \\ 84.5\%
\end{tabular} & \begin{tabular}{l} CKPlus \\ SAIT \\ SAIT 2 \\ Internet \end{tabular} & No \\ \cline{1-4}

Yu\cite{yu2015} & 61.29\% & \begin{tabular}{l} SFEW 2.0 \end{tabular} & No \\ \cline{1-4}

Ghosh\cite{ghosh2015} & \begin{tabular}{l} NA \\ 84.6\% \\ 75.8\% \\ \end{tabular} & \begin{tabular}{l} CKPlus \\ DISFA \\ BP4D \\ \end{tabular} & No \\ \cline{1-4}

Mollahosseini\cite{Mollahosseini2015} & \begin{tabular}{l} 94.7\% \\ 77.6\% \\ 55\% \\ 76.7\% \\ 47.7\% \\ 93.2\% \\ 66.4\% \end{tabular} & \begin{tabular}{l} MultiPIE\\ MMI\\ DISFA \\ FERA \\ SFEW\\ CKPlus \\ FER2013 \end{tabular} & No  \\ \cline{1-4}

EmotiW\cite{levi2015} & 54.56\% & \begin{tabular}{c}SFEW \end{tabular} & Yes  \\ \cline{1-4}

Kim\cite{Kim2015} & 61.6\% & \begin{tabular}{c}SFEW\end{tabular} & No \\ \cline{1-4}

Spiers\cite{Spiers2016} & \multicolumn{1}{c|}{\begin{tabular}{c}77.3\%\\ 89.7\%\end{tabular}} & \multicolumn{1}{l|}{\begin{tabular}{l} CKPlus\\ AM-FED\end{tabular}} & \multicolumn{1}{c|}{Yes}   \\ \cline{1-4}

DeXpression\cite{DeXpression2016} & \multicolumn{1}{l|}{\begin{tabular}{l} 99.6\%\\ 98.63\%\end{tabular}} & \multicolumn{1}{l|}{\begin{tabular}{l} CKPlus\\ MMI Facial \\ Expression\end{tabular}} & \multicolumn{1}{c|}{No}   \\ \cline{1-4}

Alizadeh\cite{Alizadeh2016} & 80.5\% & \begin{tabular}{l}FER2013 \end{tabular} & No  \\ \cline{1-4}

HappyNet \cite{duncan2016} & \begin{tabular}{l}90.7\%\\ 62.0\% \\ 90.9\% \end{tabular} & \begin{tabular}{l}CKPlus\\ Jaffe \\Homebrew \end{tabular} & Yes   \\ \cline{1-4}

Ruiz-Garcia\cite{Ruiz-Garcia2016} & 86.73\% & \begin{tabular}{l} KDEF \end{tabular} & No   \\ \cline{1-4}

FaceNet2ExpNet \cite{facenet2016} & \begin{tabular}{l} 96.8\% \\ 87.71\% \\ 88.9\% \\ 48.19\% \end{tabular} & \begin{tabular}{l}  CKPlus \\ Oulu-CASIA \\ TFD\\ SFEW \end{tabular} & No \\ \cline{1-4}

Laranjeira\cite{Laranjeira2017} & 90.0\% & \begin{tabular}{l} CKPlus \end{tabular} & No \\ \cline{1-4}

\end{tabular}
\caption{Summary of Related Work. Detection accuracy is as reported in the cited papers; in cases where different architectures were proposed, the accuracy of the best performing one is reported here, and in cases where the evaluation is done on the same dataset and on cross-datasets the reported accuracy is from the same dataset}
\label{tab:relatedwork}
\end{table*}

When it comes to the detection accuracy reported by the state-of-the-art methods, as depicted in Table \ref{tab:relatedwork}, we reported the highest accuracy given by authors. This means that when different architectures were used for the evaluation we reported the accuracy of the best performing one. Also, when the proposed method is evaluated on both individual dataset (by splitting it into training and testing part) and on cross-datasets (by training on one dataset and testing on the other) we reported the accuracy obtained on the individual dataset. We are aware that such approach gives over-promising accuracy of the state-of-the-art methods, but the main reasoning behind it was that many authors report detection accuracy only on the individual datasets, that is usually much higher than it would be the case when different datasets are used, so the comparison would not be fair among the approaches if we opted for a different choice.  
Even in this over-promising scenario, we can see that the accuracy varies from 48.19\% on the SFEW (dataset with images from the real environment presented in \cite{facenet2016}) up to 99.6\% on the 99.6\% on CKPlus (dataset created in lab environment presented in \cite{DeXpression2016}).  

\section{Experimental Setup}
\label{sec:setup}

In this Section we first describe the embedded systems we use for the evaluation of the power consumption and inference time. Then, we give more details on the dataset we used in order to evaluate detection accuracy of the observed methods, followed by more information on how we selected the candidates suitable for embedded environment.

\subsection{Used Embedded Devices}
\begin{figure}
\includegraphics[width=\columnwidth]{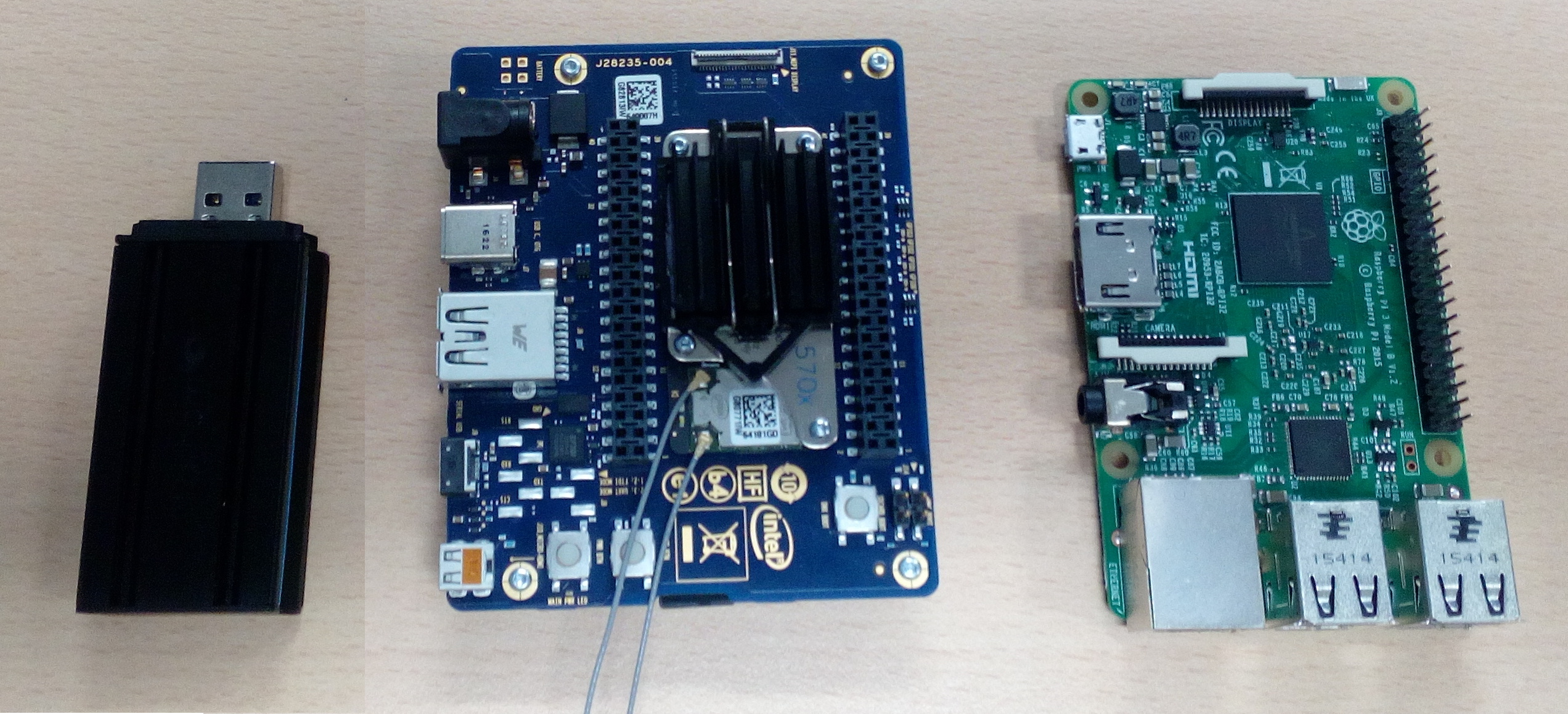}
\caption{Embedded environments used to evaluate the different CNN. From left to right: Movidius NCS, Intel Joule 570X, Raspberry Pi 3 Model B.\cite{PenaRSS2017}}
\label{fig:hwsetup}
\end{figure}

Three embedded platforms were used to evaluate the performance of the investigated face emotion recognition methods: Raspberry Pi 3 Model B \cite{raspberrypi}; Intel Joule 570X \cite{IntelJoule} and Movidius Neural Compute Stick \cite{Fathom}. These platforms, that are outlined in Figure \ref{fig:hwsetup}, are selected as they are the only ones available on the market with low power consumption and capable of performing CNN inference.

The Raspberry Pi is setup with Raspbian Jessie and the Intel Joule is setup with Ubuntu 16.04. Both of them run Tensorflow 1.0.1, Caffe with OpenBLAS with the commits id f731bc4 and 92058a7 respectively.

The Neural Compute Stick (NCS) is a low-cost and low-power USB device that enables fanless accelerated CNN inference for low-resource devices. It is based on the MA2450 Vision Processing Unit (VPU) \cite{Barry2015} developed by Movidius.

The used setup, which was kindly shared by the authors, is the same as in \cite{PenaRSS2017}. The setup uses the INA219 to measure the current/power monitor by Texas Instruments. 
The INA219 is attached to the power line of the Raspberry Pi and the Joule, then the power consumption is measured every 2ms and the average is computed over $500$ inferences. For the NCS, the INA219 is attached to the power lines of USB connection in order to measure the consumption during the inference. The sampling is the same as for the Raspberry Pi and Joule.

\subsection{Collected Dataset Description}
\label{sec:dataset}
For the collection of images and evaluation of detection accuracy, we focus on the publicly available datasets labeled with emotions, rather than AUs. 
As we can observe from Table \ref{tab:relatedwork}, the most frequently used publicly available dataset is CKPlus that contains the images created in the lab environment. SFEW dataset, that contains images created in real environment, is second frequently used. Although both datasets are publicly available, SFEW is restricted to the academic use only, so in our scenario of industrial research was not possible to use it. Thus, we collected images from CKPlus, and in addition to it we also collected data from JAFFE dataset and FER2013 datasets; the former we used in the evaluation, while the later were not used because the image size of 48x48 was too small for the most of the analyzed methods (that used images of 224x224). 
 
In order to compensate the lack of publicly available representative data for non-exaggerated face emotion expression
we collected the dataset, that we call \textit{Custom} in the remaining part of this paper.
In order to preserve the privacy of our subjects we opted to keep the dataset private. However, the main motivation behind its creation was in any case an evaluation of the detection accuracy of the existing methods on the data representative for usage in real systems and on non-exaggerated emotions, rather than the contribution in the release of a new dataset. 

The dataset was created by using a script similar to \textit{gather\_training \_data.py}, publicly available and released by the authors of HappyNet \cite{duncan2016}.
As suggested by the authors of HappyNet \cite{duncan2016}, in order to increase the number of available images for the training part and to make the detection method more robust, the images are created by adding jitter and small horizontal and vertical movements to the originally collected images. 

All collected images are RGB of size 224x224.
While the script provided by the authors is intended to show emotions in random manner, in the script we used, in order to be sure that the dataset is balanced we prompt user to express $10$ times each of seven basic emotions. 
This experiments were performed with 12 subjects, eight men and four women.
Each of the subject was prompted to show each of the seven emotions in a realistic, non-exaggerated manner, and to record it once he or she was ready.
Also, for each of the emotions the subject could choose to skip it, in case it was too difficult to express it realistically.
Additionally, participants were encouraged to record emotions in various manners. Namely, where applicable, with and without 
glasses, and/or with and without bangs.
In order to come as close as possible to the realistic scenario, no particular lighting was used for the image collection.

In Table \ref{tab:used-dataset} we summarize the used dataset. The images from these datasets were split into training, testing and validation part. 

\begin{table}[h]
\centering
\resizebox{0.4\textwidth}{!}{%
\begin{tabular}{llll}
\hline
\multicolumn{1}{|l|}{\textbf{Dataset}} & \multicolumn{1}{l|}{\textbf{No of Images}} & \multicolumn{1}{l|}{\textbf{Type}} & \multicolumn{1}{l|}{\textbf{Image Size}} \\ \hline
\multicolumn{1}{|l|}{\textbf{CKPlus}}       & \multicolumn{1}{l|}{902}                   & \multicolumn{1}{l|}{RGB}                     & \multicolumn{1}{l|}{640x480}             \\ \hline
\multicolumn{1}{|l|}{\textbf{Jaffe}}        & \multicolumn{1}{l|}{213}                   & \multicolumn{1}{l|}{Grey}                    & \multicolumn{1}{l|}{255x256}             \\ \hline
\multicolumn{1}{|l|}{\textbf{Custom}}       & \multicolumn{1}{l|}{9836}                  & \multicolumn{1}{l|}{RGB}                     & \multicolumn{1}{l|}{224x224}             \\ \hline
\end{tabular}%
}
\caption{Characteristics of the used dataset.}
\label{tab:used-dataset}
\end{table}

The used dataset consists of of $8800$ images used to train the system and the test set of $1035$ used to test its performance. 
Validation was performed on Jaffe complete dataset and CKPlus complete datasets (thus images seen by the network in either training or testing phase) and finally on Custom validation dataset containing $1116$ completely new images (thus previously never seen by the network).
We perform such separation because Custom dataset contains much more images than other two, thus allowing us to omit some of the images in the learning phase of the network development, and use them later. 

\subsection{Selection of Potentially Suitable Models}

In order to understand up to which point the existing face emotion detection methods are suitable for embedded systems and to measure in addition to detection accuracy, also power consumption and inference time, in this Section we identify those methods that are on one side accurate enough and on the other appear as potentially good candidates for embedded environment. 

According to our results on the reported detection accuracy outlined in Table \ref{tab:relatedwork}, the most accurate detection of 99.6\% was obtained using DeXpression \cite{DeXpression2016} on CKPlus dataset. In addition to reported high accuracy, the authors also report that the proposed architecture is envisioned and suitable for real world applications. Due to these two factors we consider it as a potentially good candidate for the further evaluation in our scenario. 

Second most accurate proposed architecture was from Song\cite{song2014}, where the accuracy of 99.2\% was obtained on CKPlus dataset. In the proposed architecture the CNN part is envisioned to run on a cloud based infrastructure and only report the classification outcome to the embedded device, so we did not take this approach in our evaluation.

Further candidate, that was on one side accurate and on the other proposed by authors as suitable for real-time application was HappyNet \cite{duncan2016}. This architecture relies on EmotiW architecture, introduced in \cite{levi2015}, due to which we also consider EmotiW as a suitable candidate for the analysis we perform. 

Other candidates from Table \ref{tab:relatedwork} were not considered, either because the reported detection accuracy was not very high comparing to the other methods or because the authors do not envision (or at least mention) the suitability of the proposed architectures for the embedded environment. 

Having in mind the aforementioned, in this work we implement and report detection accuracy, power consumption and inference time for following state-of-the-art networks: Dexpression \cite{DeXpression2016}, HappyNet \cite{duncan2016}, EmotiwVGG\_S\_rgb \cite{levi2015},  and nViso\cite{EoT}.

While all the other networks are described in research papers, nViso is proposed for face emotion detection task as a part of the Eyes of Things project \cite{EoT},  and its caffemodel is provided to us from the project participants. As input it uses grey scale images of size 50x50.

In case of other methods, for Dexpression \cite{DeXpression2016}, no source files were available, and it was replicated by following description given in the paper and using the dataset described in Section \ref{sec:dataset}. 
In case of HappyNet the files necessary to create caffemodel were released (training and solver configurations) that allowed us to replicate the network. However, the authors did not release the Homebrewed dataset that they used, so again we used the one described in Section \ref{sec:dataset}. 
Finally, caffemodel for EmotiwVGG\_S\_rgb \cite{levi2015} was publicly available and in our evaluation we used it in the format released by authors.

\section{Results and Discussion}
\label{sec:results}
In order to estimate their detection performance in real-time scenario, all the investigated networks are evaluated for their accuracy on the dataset described in Section \ref{sec:dataset} and for their power consumption, energy and inference time on Movidius NCS, Pi, and Joule, as described in Section \ref{sec:setup}. 

The obtained accuracy of the evaluated models is shown in Table \ref{tab:accuracy-overall}. 
As we can see from Table \ref{tab:accuracy-overall} the overall obtained accuracy for real-time like scenario is relatively low, especially comparing to the reported accuracy from Table \ref{tab:relatedwork}. 
We believe that this is the case since in our dataset we used non-exaggerated emotions, so as emotions from all seven categories, which was shown in previous work to significantly decrease detection accuracy \cite{duncan2016}, \cite{Alizadeh2016}. 
While some emotions could be avoided, that are known to be difficult to detect, i.e. fear, or images with exaggerated emotions only can be used, it was our decision to perform the evaluation in exactly such a way in order to obtain realistic estimation of the potential of current face emotion recognition methods.
Interestingly enough, in case of fear, all $12$ subjects from which we collected data, expressed difficulties in replicating this emotion correctly. Similar situation was also for disgust and sadness. Even if this was the case, we included those images in the dataset as they were, namely with the label of the requested emotion.  
Further observation related to the accuracy, that can be seen from Table \ref{tab:accuracy-overall}, is that different network perform differently on the observed three datasets. Namely, HappyNet\cite{duncan2016} and Dexpression \cite{DeXpression2016} perform well on Jaffe dataset, that is having exaggerated emotions posed by actors.
While for different networks accuracy varies depending on a dataset, nViso\cite{EoT} appears to be a robust face emotion detector, that performs reasonable well on all three datasets, and outperforms other on Custom dataset.
In case of HappyNet, we see that the detection accuracy was significantly decreased, even though the experiments are replicated using the code provided by the authors themselves. We believe that this is the case due to the fact that we used a different set of images for training, testing and validation of the model (since it was not released by authors), and due to the fact that the performance of this approach degrades significantly when the emotions are not exaggerated, when lighting is not good, and when camera is not at the eye level, as also stated by the authors at \cite{duncan2016}. 

\begin{table}[h]

\begin{tabular}{|l|c|c|c|}
\hline
\multirow{2}{*}{\textbf{Network}} & \multicolumn{3}{c|}{\textbf{Accuracy} (\%)}               \\ \cline{2-4} 
                                  & \textbf{CKPlus} & \textbf{Jaffe} & \textbf{Custom} \\ \hline
nViso \cite{EoT}                            & 21.95           & 26.76          & 23.5                    \\ \hline
HappyNet \cite{duncan2016}                         & 30.71           & 53.52          & 14.87                   \\ \hline
Dexpression \cite{DeXpression2016}                       & 23.05           & 45.07          & 14.24                   \\ \hline
EmotiW\_rgb \cite{levi2015}                       & 14.97           & 15.49          & 13.53                   \\ \hline

\end{tabular}%
\centering
\caption{Accuracy obtained for the benchmarked networks on different datasets.}
\label{tab:accuracy-overall}
\end{table}

\begin{figure*}
\centering
\includegraphics[width=0.85\textwidth]{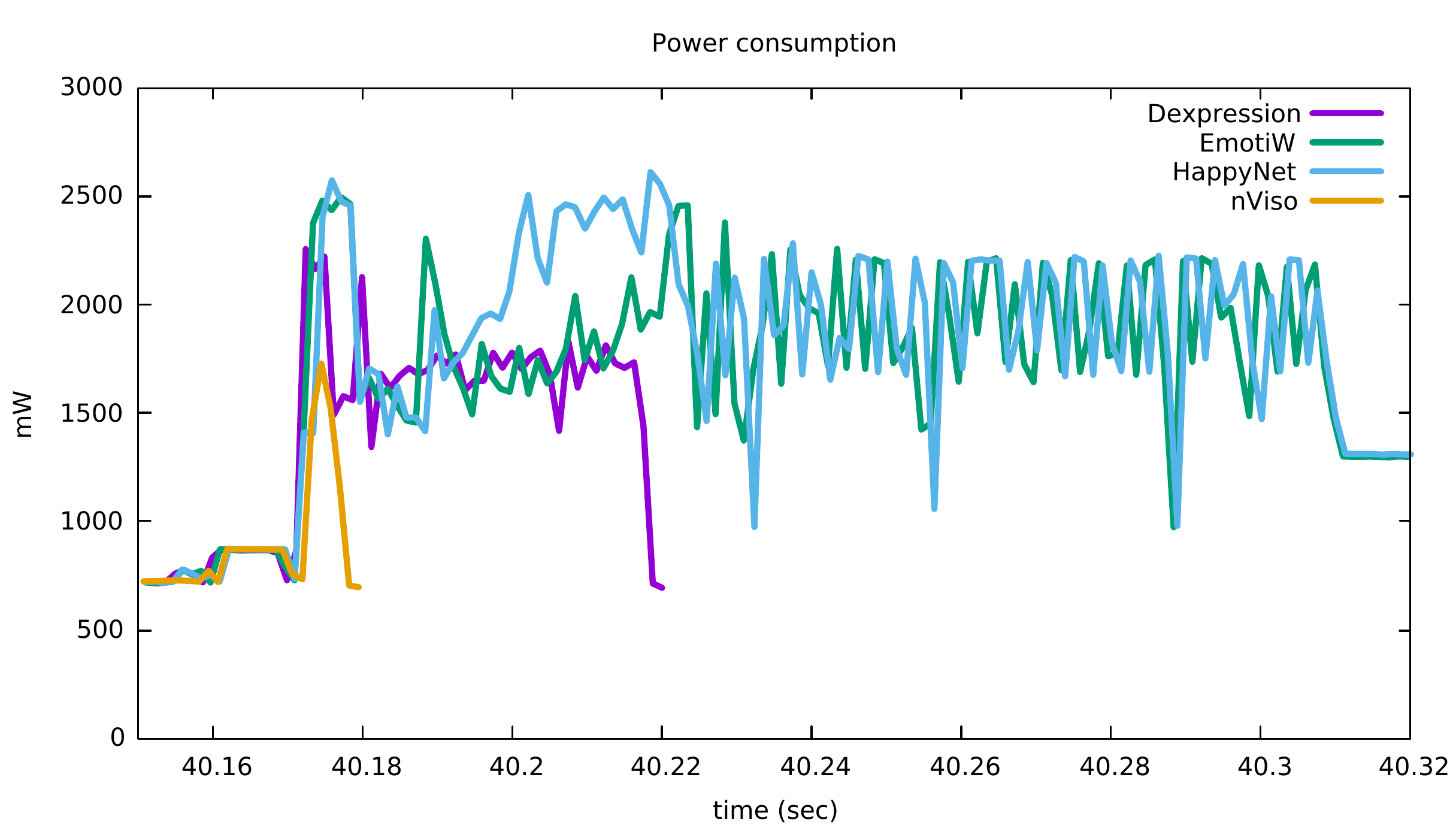}
\caption{Power profile of a single forward pass of the evaluated networks on the Movidius NCS.}
\label{fig:dexpression-ncs}
\end{figure*}

Related to the obtained results in terms of power consumption and inference, in Figure \ref{fig:dexpression-ncs} we give an overview of the outputs for a single forward pass for the observed networks on the Movidius NCS.
As it can be seen from the Figure \ref{fig:dexpression-ncs} the lowest power consumption is in case of nViso.
In our opinion, this was expected having in mind that as its input nViso uses greyscale images of size 50x50 while the other methods use RGB images of size 224x224.
In Table \ref{tab:power_usage}, we report the average power, time per forward pass and consequently energy per forward pass for three considered embedded devices and for other evaluated networks that use the same size input.
According to the results, Dexpression network \cite{DeXpression2016} is more efficient than other methods in terms of energy consumption and inference time for all considered embedded systems.  
HappyNet\cite{duncan2016} and EmotiwVGG\_S\_rgb\cite{levi2015} have very similar consumption, which is expected having in mind that HappyNet was developed using EmotiwVGG\_S\_rgb as a starting point and by modifying its last three fully connected layers.  

\begin{table*}
\centering

\begin{tabular}{|l|l|l|l|l|l|l|l|l|l|}
\hline
 & \multicolumn{9}{|c|}{\textbf{Framework}} \\
\hline
 & \multicolumn{3}{|c|}{\textbf{NCS}}  & \multicolumn{3}{|c|}{\textbf{Joule}} &  \multicolumn{3}{|c|}{\textbf{Pi}}  \\ \hline
 \textbf{Network Name} & \textbf{P (mW)} & \textbf{E (mJ)} & \textbf{T (ms)} &\textbf{P (mW)} & \textbf{E (mJ)} & \textbf{T (ms)} & \textbf{P (mW)} & \textbf{E (mJ)} & \textbf{T (ms)} \\ \hline
nViso \cite{EoT}                & 1514                                                           & 7                                                                   & 4.5                 & 4970                 & 31                        & 6                   & 730                  & 54                        & 74                  \\ \hline
HappyNet \cite{duncan2016}               & 1692                                                           & 413                                                                 & 244                 & 5143                 & 2823                      & 549                 & 699                  & 5965                      & 8528                \\ \hline
Dexpression \cite{DeXpression2016}           & 1697                                                           & 80                                                                  & 47                  & 4910                 & 832                       & 170                 & 679                  & 1262                      & 1858                \\ \hline
EmotiW\_rgb \cite{levi2015}                & 1682                                                           & 410                                                                 & 244                 & 5146                 & 2834                      & 550                 & 714                  & 6097                      & 8534                \\ \hline

\end{tabular}
\caption{Average power (P), energy consumption (E) and time (T) per inference obtained by running the network on the testing platforms: Movidius NCS, Intel Joule and Raspberry Pi 3 Model B. The Caffe framework \cite{jia2014caffe} is used for running the CNNs on the Intel Joule and Raspberry Pi.}
\label{tab:power_usage}
\end{table*}

Having in mind aforementioned observations and the obtained results in detection accuracy reported in Table \ref{tab:accuracy-overall} and power consumption and inference time reported in Table \ref{tab:power_usage}, we can summarize the encountered situation in using face emotion recognition methods for embedded environment in the following way:
\begin{itemize}
    \item When it comes to the detection accuracy, once the used dataset is with non-exaggerated emotions and without specific experimental setup, it significantly drops from what is reported in the state of the art (where for the evaluation mostly datasets with posed emotions were used) showing us that further improvements in this direction are still needed.
    \item Related to the inference time, nViso and Dexpression networks are the only ones able to run in less than 50ms. This is equivalent at a frame rate of 50 fps. Depending on the application, this can be taken as real time.
    \item Table \ref{tab:power_usage} shows that the NCS is the most energy efficient device. This is as the Myriad2 is designed to perform high number of operations on a low-power manner using the SHAVE cores. The Intel Joule and Raspberry Pi contain a general purpose CPU, therefore they are not optimized for high-performance low-power applications.
    
\end{itemize}

Based on the performed experiments and obtained results, it is our observation that current research challenges and encountered obstacles for the further adoption of the face emotion recognition methods for embedded systems are as follows:  

\begin{itemize}
    \item Analysis of grey images and use of smaller size image as input, as it is done in nViso\cite{EoT}, is still more suitable for embedded devices than RGB images and bigger image sizes, which might be a limiting factor for further accuracy improvement.
    
    \item Although different datasets are already proposed in the literature, due to specific conditions under the images are collected, they are still not representative enough for the estimation of detection accuracy of the proposed methods in real-time scenarios, and that we still lack datasets in this domain both in terms of quality and in the number of available images.
    
    \item Inference time and energy efficiency are the most critical aspects for the adoption in real-life embedded scenario, specially for battery powered applications. Out of the considered methods, only two are suitable for performing inference in a short time using a low amount of power.
    
    \item While in the state-of-the-art methods most of the results are reported using the same dataset for the evaluation of detection performance, we lack information on how the methods trained on one dataset perform on the other datasets, since this might be the common scenario of use of the proposed face emotion detection systems if adopted in real-time. 
    
\end{itemize}

\section{Conclusions}
High accuracy of face emotion detection techniques is prerequisite for adoption of these systems in real scenarios. However, low power consumption and short inference time, that are often not reported by researchers when designing and proposing such systems, are equally important. 
In order to enable using also these parameters in the evaluation of face emotion recognition methods we implemented, those existing methods that appear as the most promising candidates for real-time usage and, in this paper, reported their power consumption, energy and inference time.
Our results show that in most cases these are limiting factors for real-time usage of the developed systems.
We do hope that these findings will motivate the more comprehensive comparison between future research in this field and existing state-of-the-art methods and will empower researchers to compare their work with respect to all three metrics that matter: accuracy, power consumption and inference time. 

\balance
\bibliographystyle{plain}
\bibliography{main.bib}

\begin{thebibliography}{10}

\bibitem{raspberrypi}
{Raspberry Pi}.
\newblock Online: https://www.raspberrypi.org/.

\bibitem{fer2013}
{Challenges in Representation Learning: Facial Expression Recognition
  Challenge}, 2013.
\newblock Online:
  https://www.kaggle.com/c/challenges-in-representation-learning-facial-expression-recognition-challenge/data.

\bibitem{EoT}
{Eyes of Things}, 2018.
\newblock Online: http://eyesofthings.eu.

\bibitem{IntelJoule}
{Intel Joule}, 2018.
\newblock Online: https://software.intel.com/en-us/intel-joule-getting-started.

\bibitem{Fathom}
{Software Development Kit for the Neural Compute Stick}, 2018.
\newblock Online: https://github.com/movidius/ncsdk/.

\bibitem{Alizadeh2016}
Shima Alizadeh and Azar Fazel.
\newblock {Convolutional Neural Networks for Facial Expression Recognition}.
\newblock Online: http://cs231n.stanford.edu/reports/2016/pdfs/005{\_}
  Report.pdf.

\bibitem{Barry2015}
Brendan Barry, Cormac Brick, Fergal Connor, David Donohoe, David Moloney,
  Richard Richmond, Martin O'Riordan, and Vasile Toma.
\newblock {Always-on Vision Processing Unit for Mobile Applications}.
\newblock {\em IEEE Micro}, 35(2):56--66, mar 2015.
\newblock Online: http://ieeexplore.ieee.org/document/7024073/.

\bibitem{DeXpression2016}
Peter Burkert, Felix Trier, Muhammad~Zeshan Afzal, Andreas Dengel, and Marcus
  Liwicki.
\newblock {DeXpression: Deep Convolutional Neural Network for Expression
  Recognition}.
\newblock Online: https://arxiv.org/pdf/1509.05371.pdf.

\bibitem{Deniz2017}
Oscar Deniz, Noelia Vallez, Jose Espinosa-Aranda, Jose Rico-Saavedra, Javier
  Parra-Patino, Gloria Bueno, David Moloney, Alireza Dehghani, Aubrey Dunne,
  Alain Pagani, Stephan Krauss, Ruben Reiser, Martin Waeny, Matteo Sorci, Tim
  Llewellynn, Christian Fedorczak, Thierry Larmoire, Marco Herbst, Andre
  Seirafi, and Kasra Seirafi.
\newblock {Eyes of Things}.
\newblock {\em Sensors}, 17(6):1173, may 2017.
\newblock Online: http://www.mdpi.com/1424-8220/17/5/1173.

\bibitem{sfew}
A.~Dhall, R.~Goecke, S.~Lucey, and T.~Gedeon.
\newblock Static facial expression analysis in tough conditions: Data,
  evaluation protocol and benchmark.
\newblock In {\em 2011 IEEE International Conference on Computer Vision
  Workshops (ICCV Workshops)}, pages 2106--2112, Nov 2011.

\bibitem{facenet2016}
Hui Ding, Shaohua~Kevin Zhou, and Rama Chellappa.
\newblock {FaceNet2ExpNet: Regularizing a Deep Face Recognition Net for
  Expression Recognition}.
\newblock {\em CoRR}, abs/1609.06591, 2016.

\bibitem{duncan2016}
Dan Duncan, Gautam Shine, and Chris English.
\newblock {Facial Emotion Recognition in Real Time}.
\newblock Online: http://cs231n.stanford.edu/reports/2016/pdfs/022{\_}
  Report.pdf.

\bibitem{ghosh2015}
S.~Ghosh, E.~Laksana, S.~Scherer, and L.~P. Morency.
\newblock A multi-label convolutional neural network approach to cross-domain
  action unit detection.
\newblock In {\em 2015 International Conference on Affective Computing and
  Intelligent Interaction (ACII)}, pages 609--615, Sept 2015.

\bibitem{pie2}
Ralph Gross, Iain Matthews, Jeff Cohn, Takeo Kanade, and Simon Baker.
\newblock Multi-pie.
\newblock {\em Proceedings of the International Conference on Automatic Face
  and Gesture Recognition. International Conference on Automatic Face and
  Gesture Recognition}, 28(5):807--813, 05 2010.

\bibitem{pie1}
Ralph Gross, Iain Matthews, Jeffrey Cohn, Takeo Kanade, and Simon Baker.
\newblock Multi-pie.
\newblock {\em Image Vision Comput.}, 28(5):807--813, May 2010.
\newblock Online: http://dx.doi.org/10.1016/j.imavis.2009.08.002.

\bibitem{jia2014caffe}
Yangqing Jia, Evan Shelhamer, Jeff Donahue, Sergey Karayev, Jonathan Long, Ross
  Girshick, Sergio Guadarrama, and Trevor Darrell.
\newblock Caffe: Convolutional architecture for fast feature embedding.
\newblock {\em arXiv preprint arXiv:1408.5093}, 2014.

\bibitem{ck}
Takeo Kanade, Yingli Tian, and Jeffrey~F. Cohn.
\newblock Comprehensive database for facial expression analysis.
\newblock In {\em Proceedings of the Fourth IEEE International Conference on
  Automatic Face and Gesture Recognition 2000}, FG '00, pages 46--, Washington,
  DC, USA, 2000. IEEE Computer Society.
\newblock Online: http://dl.acm.org/citation.cfm?id=795661.796155.

\bibitem{Kim2015}
Bo-Kyeong Kim, Hwaran Lee, Jihyeon Roh, and Soo-Young Lee.
\newblock Hierarchical committee of deep cnns with exponentially-weighted
  decision fusion for static facial expression recognition.
\newblock In {\em Proceedings of the 2015 ACM on International Conference on
  Multimodal Interaction}, ICMI '15, pages 427--434, New York, NY, USA, 2015.
  ACM.
\newblock Online: http://doi.acm.org/10.1145/2818346.2830590.

\bibitem{Laranjeira2017}
Ana Laranjeira, Xavier Fraz{\~a}o, Andr{\'e} Pimentel, and Bernardete Ribeiro.
\newblock {\em How Deep Can We Rely on Emotion Recognition}, pages 511--520.
\newblock Springer International Publishing, 2017.

\bibitem{levi2015}
Gil Levi and Tal Hassner.
\newblock Emotion recognition in the wild via convolutional neural networks and
  mapped binary patterns.
\newblock In {\em Proc. ACM International Conference on Multimodal Interaction
  (ICMI)}, November 2015.
\newblock Online: http://www.openu.ac.il/home/hassner/projects/cnn\_emo-tions.

\bibitem{ckplus}
Patrick Lucey, Jeffrey~F. Cohn, Takeo Kanade, Jason Saragih, Zara Ambadar, and
  Iain Matthews.
\newblock The extended cohn-kanade dataset (ck+): A complete dataset for action
  unit and emotion-specified expression.

\bibitem{kdef}
D.~Lundqvist, A.~Flykt, and A.~Ohman.
\newblock {The Karolinska Directed Emotional Faces - KDEF}.
\newblock CD ROM from Department of Clinical Neuroscience, Psychology section,
  Karolinska Institutet, 1998.

\bibitem{jaffe1}
M.~Lyons, S.~Akamatsu, M.~Kamachi, and J.~Gyoba.
\newblock Coding facial expressions with gabor wavelets.
\newblock In {\em Proceedings Third IEEE International Conference on Automatic
  Face and Gesture Recognition}, pages 200--205, Apr 1998.

\bibitem{jaffe2}
Michael~J. Lyons, Julien Budynek, and Shigeru Akamatsu.
\newblock Automatic classification of single facial images.
\newblock {\em IEEE Trans. Pattern Anal. Mach. Intell.}, 21(12):1357--1362,
  December 1999.
\newblock Online: http://dx.doi.org/10.1109/34.817413.

\bibitem{disfa}
S.~M. Mavadati, M.~H. Mahoor, K.~Bartlett, P.~Trinh, and J.~F. Cohn.
\newblock Disfa: A spontaneous facial action intensity database.
\newblock {\em IEEE Transactions on Affective Computing}, 4(2):151--160, April
  2013.

\bibitem{Mollahosseini2015}
Ali Mollahosseini, David Chan, and Mohammad~H Mahoor.
\newblock {Going Deeper in Facial Expression Recognition using Deep Neural
  Networks}, 2015.
\newblock Online: https://arxiv.org/pdf/1511.04110.pdf.

\bibitem{mmi}
M.~Pantic, M.~Valstar, R.~Rademaker, and L.~Maat.
\newblock Web-based database for facial expression analysis.
\newblock In {\em 2005 IEEE International Conference on Multimedia and Expo},
  pages 5 pp.--, July 2005.

\bibitem{PenaRSS2017}
Dexmont Pena, Andrew Forembski, Xiaofan Xu, and David Moloney.
\newblock {Benchmarking of CNNs for Low-Cost, Low-Power Robotics Applications}.
\newblock In {\em RSS 2017 Workshop: New Frontier for Deep Learning in
  Robotics}, 2017.

\bibitem{Ruiz-Garcia2016}
Ariel Ruiz-Garcia, Mark Elshaw, Abdulrahman Altahhan, and Vasile Palade.
\newblock {Deep Learning for Emotion Recognition in Faces}.
\newblock pages 38--46. Springer, Cham, 2016.
\newblock Online: http://link.springer.com/10.1007/978-3-319-44781-0{\_}5.

\bibitem{song2014}
I.~Song, H.~J. Kim, and P.~B. Jeon.
\newblock Deep learning for real-time robust facial expression recognition on a
  smartphone.
\newblock In {\em 2014 IEEE International Conference on Consumer Electronics
  (ICCE)}, pages 564--567, Jan 2014.

\bibitem{Spiers2016}
Daniel~Llatas Spiers.
\newblock {Facial emotion detection using deep learning}.
\newblock 2016.
\newblock Online:
  http://uu.diva-portal.org/smash/get/diva2:952138/FULLTEXT01.pdf.

\bibitem{fera}
M.~F. Valstar, T.~Almaev, J.~M. Girard, G.~McKeown, M.~Mehu, L.~Yin, M.~Pantic,
  and J.~F. Cohn.
\newblock Fera 2015 - second facial expression recognition and analysis
  challenge.
\newblock In {\em 2015 11th IEEE International Conference and Workshops on
  Automatic Face and Gesture Recognition (FG)}, volume~06, pages 1--8, May
  2015.

\bibitem{yu2015}
Zhiding Yu and Cha Zhang.
\newblock Image based static facial expression recognition with multiple deep
  network learning.
\newblock In {\em Proceedings of the 2015 ACM on International Conference on
  Multimodal Interaction}, ICMI '15, pages 435--442, New York, NY, USA, 2015.
  ACM.
\newblock Online: http://doi.acm.org/10.1145/2818346.2830595.

\end{thebibliography}

\end{document}